\PassOptionsToPackage{pdftex,dvipsnames}{xcolor}
\PassOptionsToPackage{hidelinks}{hyperref}
\documentclass[sigconf]{acmart}
\usepackage{booktabs} 
\usepackage{tabularx}

\usepackage{amsmath}
\usepackage{amssymb}
\usepackage{wrapfig}
\usepackage{multirow}
\usepackage{multicol}
\usepackage{enumitem}
\usepackage{array}
\usepackage{color, soul}
\usepackage[dvipsnames]{xcolor}

\usepackage{setspace}

\usepackage{subcaption}

\usepackage{graphicx,array}
\graphicspath{ {fig/} }

\usepackage{xargs}
\usepackage[colorinlistoftodos,prependcaption,textsize=tiny]{todonotes}
\newcommandx{\unsure}[2][1=]{\todo[linecolor=red,backgroundcolor=red!25,bordercolor=red,#1]{#2}}
\newcommandx{\change}[2][1=]{\todo[linecolor=blue,backgroundcolor=blue!25,bordercolor=blue,#1]{#2}}
\newcommandx{\info}[2][1=]{\todo[linecolor=OliveGreen,backgroundcolor=OliveGreen!25,bordercolor=OliveGreen,#1]{#2}}
\newcommandx{\improvement}[2][1=]{\todo[linecolor=Plum,backgroundcolor=Plum!25,bordercolor=Plum,#1]{#2}}
\newcommandx{\thiswillnotshow}[2][1=]{\todo[disable,#1]{#2}}
\setcopyright{acmcopyright}

\copyrightyear{2017}
\acmYear{2017}
\setcopyright{acmcopyright}
\acmConference{SIGIR '17}{August 07-11, 2017}{Shinjuku, Tokyo, Japan}\acmPrice{15.00}\acmDOI{http://dx.doi.org/10.1145/3077136.3080740}
\acmISBN{978-1-4503-5022-8/17/08}

\begin{document}

\abovedisplayskip=0pt
\abovedisplayshortskip=0pt
\belowdisplayskip=0pt
\belowdisplayshortskip=0pt

\title{Contextualizing Citations for Scientific Summarization using Word Embeddings and Domain Knowledge}


\author{Arman Cohan}
\affiliation{%
  \institution{Information Retrieval Lab, Dept. of Computer Science}
  \streetaddress{Georgetown University}
}
\email{arman@ir.cs.georgetown.edu}

\author{Nazli Goharian}
\affiliation{%
  \institution{Information Retrieval Lab, Dept. of Computer Science}
  \streetaddress{Georgetown University}
}
\email{nazli@ir.cs.georgetown.edu}








\newcommand\I{\textdagger}
\newcommand\II{\textdaggerdbl}

\newcommand\B[1]{\textbf{#1}}
\newcommand\SC[1]{\textsc{#1}}

\begin{abstract}

  Citation texts are sometimes not very informative or in some cases inaccurate by themselves; they need the appropriate context from the referenced paper to reflect its exact contributions. To address this problem, we propose an unsupervised model that uses distributed representation of words as well as domain knowledge to extract the appropriate context from the reference paper. Evaluation results show the effectiveness of our model by significantly outperforming the state-of-the-art. We furthermore demonstrate how an effective contextualization method results in improving citation-based summarization of the scientific articles.

\end{abstract}

%
%



\keywords{Text Summarization, Scientific Text, Information Retrieval}

\maketitle


\vspace{-6pt}
\section{Introduction}
\label{sec:intro}

In scientific literature, related work is often referenced along with a short textual description regarding that work which we call \textit{citation text}.
Citation texts usually highlight certain contributions of the referenced paper and a set of citation texts to a reference paper can provide useful information about that paper. Therefore, citation texts have been previously used to enhance many downstream tasks in IR/NLP such as search and summarization (e.g. \cite{Ritchie:2008:IR,Qazvinian2008scientific,cohan-goharian:2015:EMNLP}).

While useful, citation texts might lack the appropriate context from the reference article \cite{teufel2002summarizing,de2012epistemic,Cohan2015}. For example, details of the methods, assumptions or conditions for the obtained results are often not mentioned. Furthermore, in many cases the citing author might misunderstand or misquote the referenced paper and ascribe contributions to it that are not intended in that form. Hence, sometimes the citation text is not sufficiently informative or in other cases, even inaccurate \cite{sandor2012identifying}. This problem is more serious in life sciences where accurate dissemination of knowledge has direct impact on human lives.

We present an approach for addressing such concerns by adding the appropriate context from the reference article to the citation texts.  Enriching the citation texts with relevant context from the reference paper helps the reader to better understand the context for the ideas, methods or findings stated in the citation text.

A challenge in citation contextualization is the discourse and terminology variations between the citing and the referenced authors. Hence, traditional IR models that rely on term matching for finding the relevant information are ineffective.

We propose to address this challenge by a model that utilizes word embeddings and domain specific knowledge. Specifically, our approach is a retrieval model for finding the appropriate context of citations, aimed at capturing terminology variations and paraphrasing between the citation text and its relevant reference context.

 We perform two sets of experiments to evaluate the performance of our system. First, we evaluate the relevance of extracted contexts intrinsically. Then we evaluate the effect of citation contextualization on the application of scientific summarization. Experimental results on TAC 2014 benchmark show that our approach significantly outperforms several strong baselines in extracting the relevant contexts. We furthermore, demonstrate that our contextualization models can enhance summarizing scientific articles.

\vspace{-8pt}
\section{Contextualizing citations}
\label{sec:method}

Given a citation text, our goal is to extract the most relevant context to it in the reference article. These contexts are essentially certain textual spans within the reference article. Throughout, colloquially, we refer to the citation text as query and reference spans in the reference article as documents.
Our approach extends Language Models for IR  (LM) by incorporating word embeddings and domain ontology to address shortcomings of LM for this research purpose. The goal in LM is to rank a document $d$ according to the conditional probability $p(d|q) \propto p(q|d)=\prod_{q_i\in q}p(q_i|d)$ where $q_i$ shows the tokens in the query $q$. Estimating $p(q_i|d)$ is often achieved by maximum likelihood estimate from term frequencies with some sort of smoothing. Using Dirichlet smoothing \cite{zhai2004study}, we have:
\begin{equation}\label{eq:1}
p(q_i|d)=\dfrac{f(q_i,d)+\mu\; p(q_i|\mathcal{C})}{\sum_{w\in \mathcal{V}}{f(w,d)}+\mu}
\end{equation}
where $f(q_i,d)$ shows the frequency of term $q_i$ in document $d$, $\mathcal{C}$ is the entire collection, $\mathcal{V}$ is the vocabulary and $\mu$ the Dirichlet smoothing parameter.
In the citation contextualization problem, \textit{(i)} the target reference sentences are short documents and \textit{(ii)} there exist terminology variations between the citing author and the referenced author. Hence, the citation terms usually do not appear in the documents and relying only on the frequencies of citation terms in the documents ($f(q_i,d)$) for estimating $p(q_i|d)$ yields an almost uniform smoothed distribution that is unable to decisively distinguish between the documents.

\textit{Embeddings.} Distributed representation (embedding) of a word $w$ in a field $\mathbb{F}$ is a mapping $w\to\mathbb{F}^n$ where words semantically similar to $w$ will be ideally located close to it.
Given a query $q$, we rank the documents $d$ according to the following scoring function which leverages this property:
\begin{equation}\label{eq:2}
p(q_i|d)=\dfrac{f_{sem}(q_i,d)+\mu\; p(q_i|C)}{\sum_{w\in V}{f_{sem}(w,d)}+\mu}
\end{equation}

\noindent where $f_{sem}$ is a function that measures semantic relatedness of the query term $q_i$ to the document $d$ and is defined as: $f_{sem}(q_i,d)=\sum_{d_j\in d}s(q_i,d_j)$; where $d_j$'s are document terms and $s(q_i,d_j)$ is the relatedness between the the query term and document term which is calculated by applying a similarity function to the distributed representations of $q_i$ and $d_j$. We use a transformation ($\phi$) of dot products between the unit vectors $e(q_i)$ and $e(d_j)$ corresponding to the embeddings of the terms $q_i$ and $d_j$ for the similarity function:
$
\small s(q_i,d_j)=
\begin{cases}
\phi(e(q_i).e(d_j)); & \text{if } e(q_i).e(d_j) > \tau \\
0;              & \text{otherwise}
\end{cases}
$
\begin{figure}[t]
\begin{subfigure}{0.49\linewidth}
\centering
\includegraphics[width=3.8cm]{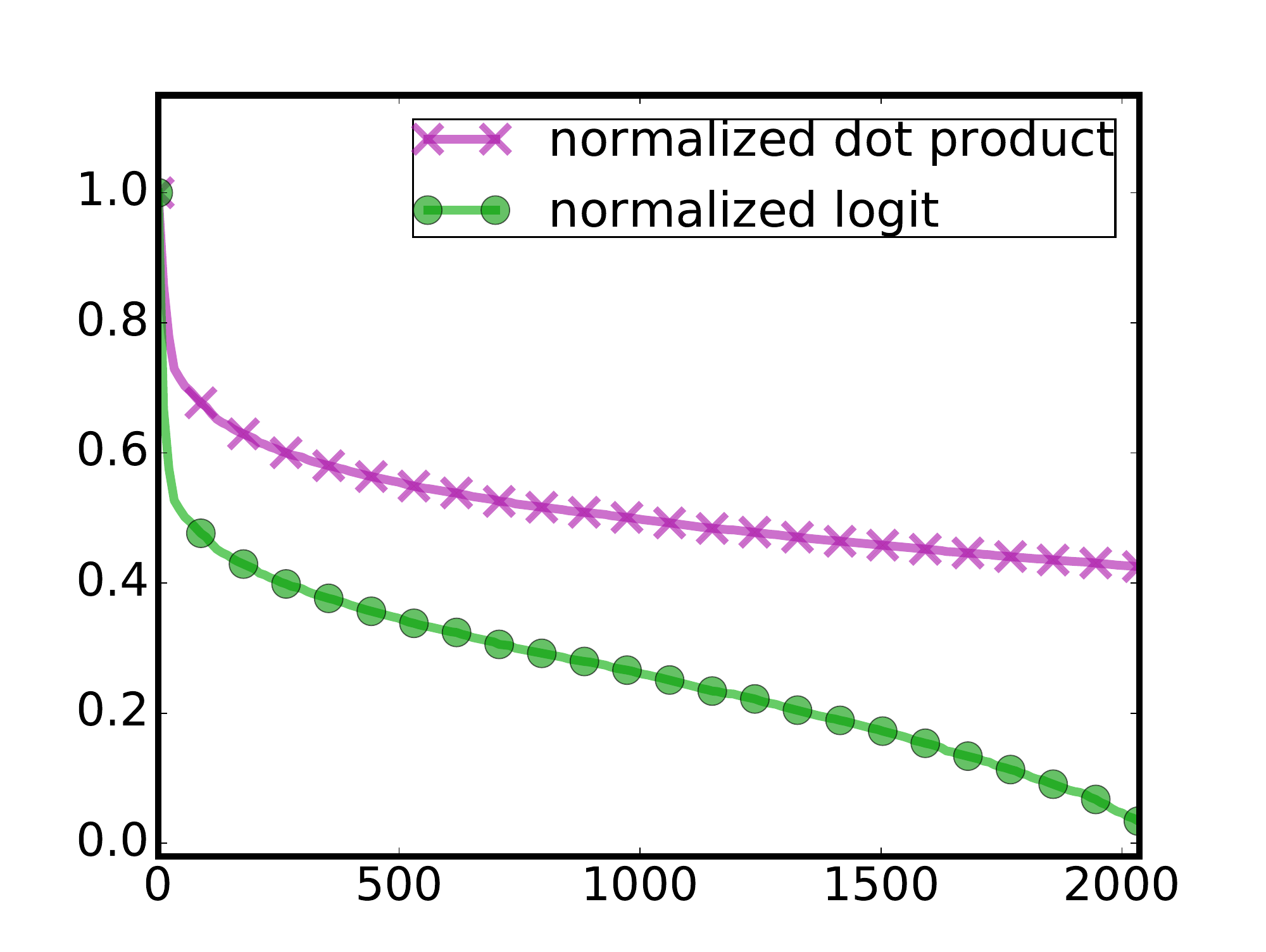}
\end{subfigure}
\begin{subfigure}{0.49\linewidth}
\centering
\includegraphics[width=3.8cm]{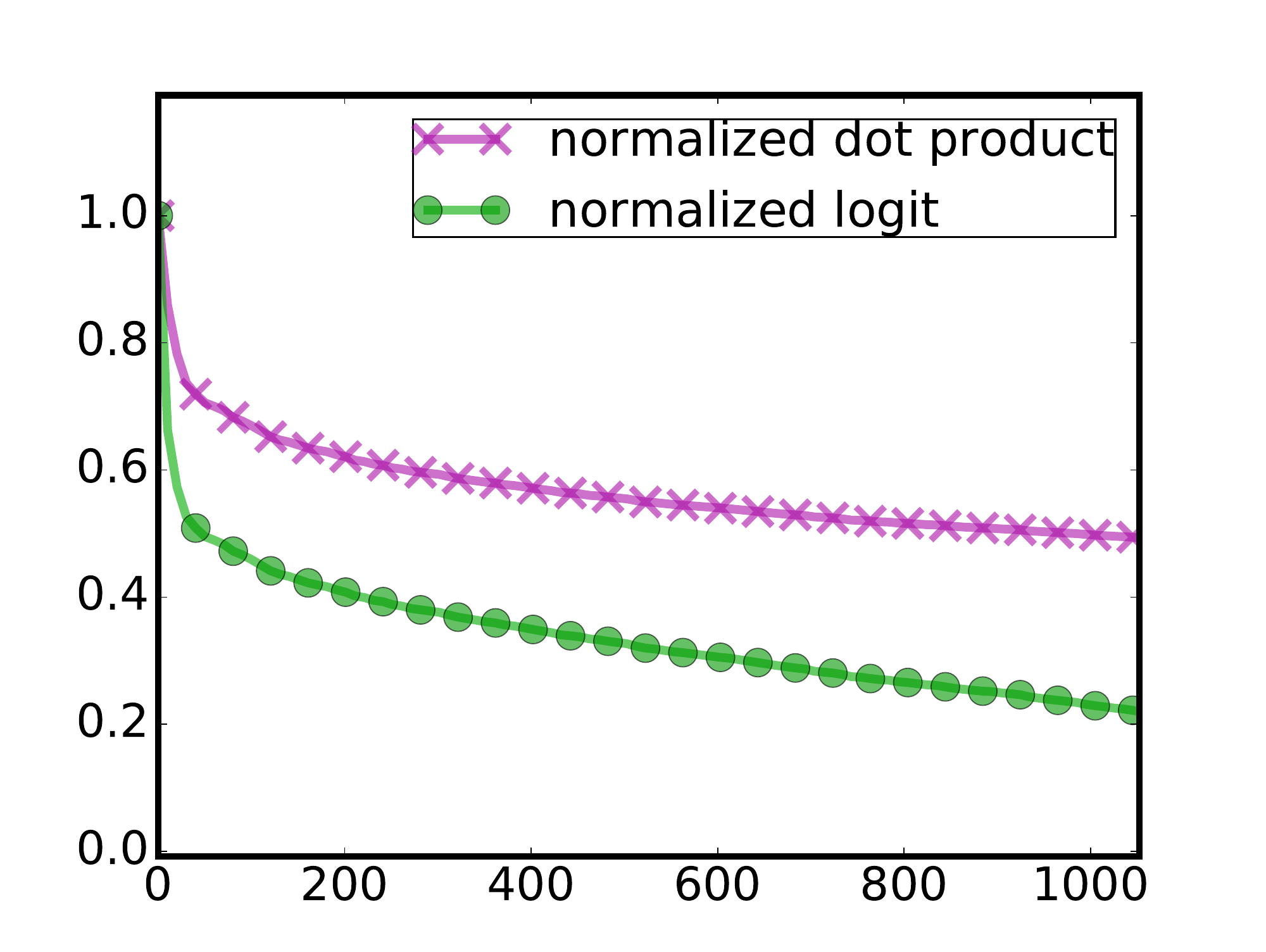}
\end{subfigure}
\vspace{-8pt}
\caption{Dot product of embeddings and its logit for a sample word and its top most similar words (top 2000 and 1000).}
\label{fig:simil}
\vspace{-18pt}
\end{figure}

We first explain the role of $\tau$ and then the reason for considering the function $\phi$ instead of raw dot product.
$\tau$ is a parameter that controls the noise introduced by less similar words. Many unrelated word vectors have non-zero similarity scores and adding them up introduces noise to the model and reduces the performance. $\tau$'s function is to set the similarity between unrelated words to zero instead of a positive number. To identify an appropriate value for $\tau$, we select a random set of words from the embedding model and calculate the average and standard deviation of pointwise absolute value of similarities between terms from these two samples. We then select the threshold $\tau$ to be two standard deviations larger than the average to only consider very high similarity values (this choice was empirically justified).

Examining term similarity values between words shows that there are many terms with high similarities associated with each term and these values are not highly discriminative. We apply a transfer function $\phi$ to the dot product $e(q_i).e(d_j)$ to dampen the effect of less similar words. In other words, we only want highly related words to have high similarity values and similarity should quickly drop as we move to less related words. We use the \textit{logit} function for $\phi$ to achieve this dampening effect:
$$\phi(x) = \log(\frac{x}{1-x})$$
Figure \ref{fig:simil} shows this effect. The purple line is the normalized dot product of a sample word with the most similar words in the model. As illustrated, the similarity score differences among top words is not very discriminative. However, applying the logit function (green line) causes the less similar words to have lower similarity values to the target word.

\vspace{-6pt}
\paragraph{Domain knowledge}

 Successful word embedding methods have previously shown to be effective in capturing syntactic and semantic relatedness between terms. These co-occurrence based models are data driven. On the other hand, domain ontologies and lexicons that are built by experts include some information that might not be captured by embedding methods \cite{hill2015simlex}. Therefore, using domain knowledge can further help the embedding based retrieval model; we incorporate it in our model in the following ways:

\noindent 1) \textit{Retrofitting}: \citet{faruqui2015retrofitting} proposed a model that uses the constraints on WordNet lexicon to modify the word vectors and pull synonymous words closer to each other.
To inject the domain knowledge in the embeddings, we apply this model on two domain specific ontologies, namely, \textsc{Mesh} and Protein Ontologies (PO)\footnote{\urlstyle{rm}\url{https://www.nlm.nih.gov/mesh/};  \hspace{6pt} \urlstyle{rm}\url{http://pir.georgetown.edu/pro/}}. We chose these two biomedical domain ontologies because they are in the same domain as the articles in the TAC dataset. \textsc{Mesh} is a broad ontology that consists of biomedical terms and PO is a more focused ontology related to biology of proteins and genes.

\noindent 2) \textit{Interpolating in the LM}: We also directly incorporate the domain knowledge in the retrieval model; we modify the LM into the following interpolated LM with parameter $\lambda$:
$$
p(q_i|d)=\lambda p_{1}(q_i|d) + (1-\lambda)p_{2}(q_i|d)
$$
\noindent where $p_1$ is estimated using Eq. \ref{eq:2} and $p_2$ is similar to $p_1$ except that we replace $f_{sem}$ with the function $f_{ont}$ which considers domain ontology in calculating similarities:
\begin{align*}
\footnotesize
f_{ont}(q_i,d){=}\sum_{d_j\in d}s_2(q_i,d_j);
~
s_2(q_i,d_j){=}
\begin{cases}
1, & \text{if } q_i{=}d_j \\
\gamma, & \text{if } q_i{\approx}d_j \\
0, & \text{o.w.}
\end{cases}
\end{align*}

where $\gamma \in [0,1]$ is a parameter and $q_i\approx d_j$ shows that there is an \textit{is-synonym} relation in ontology between $q_i$ and $d_j$\footnote{The values of the parameters $\gamma$ and $\lambda$ were selected empirically by grid search}.

\vspace{-6pt}
\section{Experiments}
\label{sec:experimental}

\textit{Data.} We use the TAC 2014 Biomedical Summarization benchmark\footnote{\urlstyle{rm}\url{http://www.nist.gov/tac/2014/BiomedSumm/}}. This dataset contains 220 scientific biomedical journal articles and 313 total citation texts where the relevant contexts for each citation text are annotated by 4 experts.

\textit{Baselines.} To our knowledge, the only published results on TAC 2014 is \cite{Cohan2015}, where the authors utilized query reformulation (QR) based on UMLS ontology. In addition to \cite{Cohan2015}, we also implement several other strong baselines to better evaluate the effectiveness of our model:  1) \textit{BM25}; 2) \textit{VSM}: Vector Space Model that was used in \cite{Cohan2015}; 3) \textit{DESM}: Dual Embedding Space Model which is a recent embedding based retrieval model \cite{mitra2016dual}; and 4) \textit{LMD-LDA}: Language modeling with LDA smoothing which is a recent extension of the LMD to also account for the latent topics \cite{jian2016simple}.
All the baseline parameters are tuned for the best performance, and the same preprocessing is applied to all the baselines and our methods.

\textit{Our methods.}
 We first report results based on training the embeddings on Wikipedia ($\mathrm{WE}_\mathrm{Wiki}$). Since TAC dataset is in biomedical domain, many of the biomedical terms might be either out-of-vocabulary or not captured in the correct context using general embeddings, therefore we also train biomedical embeddings ($\mathrm{WE}_\mathrm{Bio})$\footnote{We train biomedical embeddings on TREC Genomics 2004 and 2006 collections (both Wikipedia and Genomics embeddings were trained using gensim implementation of Word2Vec, negative sampling, window size of 5 and 300 dimensions.}. In addition, we report results for biomedical embeddings with retrofitting ($\mathrm{WE}_\mathrm{Bio}$+rtrft), as well as interpolating domain knowledge ($\mathrm{WE}_\mathrm{Bio}$+dmn)

\begin{table}[]
\centering
\small
\setlength{\tabcolsep}{1.4pt}
\renewcommand{\arraystretch}{0.65}
\caption{Results on TAC 2014 dataset. c-P, c-R, c-F: character offset Precision, Recall and F-1 scores; \textsc{Rg}: \textsc{Rouge}; c-P@K: character offset precision at K. \textdagger shows statistical significant improvement over the best baseline performance (two-tailed t-test, $p{<}0.05$). Values are percentages.\vspace{-8pt}
}
\label{tab:1}
\begin{tabular}{@{}lrrrrrrrrr@{}}
\toprule
Method                              & c-P                      & c-R                      & c-F                      & nDCG                     & \textsc{Rg}-1            & \textsc{Rg}-2            & \textsc{Rg}-3            & c-P@1                    & c-P@5                    \\ \midrule
VSM  \cite{Cohan2015}               & 20.5                     & 24.7                     & 21.2                     & 48.1                     & 49.5                     & 26.4                     & 20.0                     & 31.9                     & 26.1                     \\
BM25   & 19.5                     & 18.6                     & 17.8                     & 38.1                     & 43.6                     & 23.2                     & 16.3                     & 25.5                     & 24.2                     \\
DESM \cite{mitra2016dual}           & 20.3                     & 23.8                     & 22.3                     & 45.6                     & 50.3                     & 26.2                     & 20.6                     & 32.5                     & 26.5                     \\
LMD-LDA \cite{jian2016simple}       & 22.6                     & 24.8                     & 22.3                     & 46.0                     & 48.3                     & 26.4                     & 20.1                     & 31.4                     & 27.7                     \\
QR \cite{Cohan2015}                 & 22.2                     & 29.4                     & 23.8                     & 49.8                     & 50.6                     & 27.2                     & 21.8                     & 37.7                     & 28.1                     \\ \midrule
$\mathrm{WE}_\mathrm{Wiki}$         & 21.8                     & 28.5                     & 23.2                     & \I 52.8          & 50.0                     & 26.9                     & 20.9                     & 36.5                     & 29.9                     \\
$\mathrm{WE}_\mathrm{Bio}$          & 23.9                     & \I 31.2          & \I 25.5          & \I 57.1          & 51.9                     & \I 29.2          & \I 23.1          & \I 46.2          & \I 34.1          \\
$\mathrm{WE}_\mathrm{Bio}$+rtrft& \I 24.8          & \I \textbf{33.6} & \I 26.4          & \I 58.3          & 52.4                     & \I \textbf{30.7} & \I 24.0          & \I 55.5          & \I 34.9          \\
$\mathrm{WE}_\mathrm{Bio}$+dmn   & \I \textbf{25.4} & \I 33.0          & \I \textbf{27.0} & \I \textbf{59.8} & \I \textbf{53.0} & \I 30.6          & \I \textbf{24.4} & \I \textbf{56.1} & \I \textbf{37.1} \\ \bottomrule
\end{tabular}
\vspace{-2pt}
\end{table}

\subsection{Intrinsic Evaluation}

First, we analyze the effectiveness of our proposed approach for contextualization intrinsically. That is, we evaluate the quality of the extracted citation contexts using our contextualization methods in terms of how accurate they are with respect to human annotations.

\textit{Evaluation.} We consider the following evaluation metrics for assessing the quality of the retrieved contexts for each citation from multiple aspects:
(\textit{i}) Character offset overlaps of the retrieved contexts with human annotations in terms of precision (c-P), recall (c-R) and F-score (c-F). These are the recommended metrics for the task per TAC\footnote{https://tac.nist.gov/2014/BiomedSumm/guidelines.html}.
(\textit{ii}) nDCG: we treat any partial overlaps with the gold standard as a correct context and then calculate the nDCG scores.
(\textit{iii}) \textsc{Rouge}-N scores:
To also consider the content similarity of the retrieved contexts with the gold standard, we calculate the \textsc{Rouge} scores between them.
(\textit{iv}) Character precision at $K$ (c-P@K): Since we are usually interested in the top retrieved spans, we consider character offset precision only for the top $K$ spans and we denote it with ``c-P@K''.

\textit{Results.} The results of intrinsic evaluation of contextualization are presented in Table \ref{tab:1}. Our models (last 4 rows of table \ref{tab:1}) achieve significant improvements over the baselines consistently across most of the metrics. This shows the effectiveness of our models viewed from different aspects in comparison with the baselines. The best baseline performance is the query reformulation (QR) method by \cite{Cohan2015} which improves over other baselines.

\begin{table}[]
  \caption{Top relevant words to the word ``expression'' according to embeddings trained on Wikipedia vs. Genomics.\vspace{-8pt}}
  \label{tab:emb}
\centering
\small
\renewcommand*{\arraystretch}{0.60}
\begin{tabular}{@{}lr@{}} \toprule
General (Wikipedia)      & Biomedical (Genomics)       \\ \midrule
interpretation & upregulation   \\
sense          & mrna           \\
emotion        & induction      \\
function       & protein        \\
intension      & abundance      \\
manifestation  & gene           \\
expressive     & downregulation \\
\bottomrule
\end{tabular}
\end{table}

We observe that using general domain embeddings does not provide much advantage in comparison with the best baseline (compare $\mathrm{WE}_{wiki}$ and QR in the Table). However, using the domain specific embeddings ($\mathrm{WE}_{Bio}$) results in 10\% c-F improvement over the best baseline. This is expected since word relations in the biomedical context are better captured with biomedical embeddings. In Table \ref{tab:emb} an illustrative word ``expression'' gives better intuition why is that the case. As shown, using general embeddings (left column in the table), the most similar words to ``expression'' are those related to the general meaning of it. However, many of these related words are not relevant in the biomedical context. In the biomedical context, ``expression'' refers to ``the appearance in a phenotype attributed to a particular gene''. As shown on the right column, the domain specific embeddings (Bio) trained on genomics data are able to capture this meaning. This further confirms the inferior performance of the out-of-domain word embeddings in capturing correct word-level semantics \cite{nallapati2016classify}. Last two rows in Table \ref{tab:1} show incorporating the domain knowledge in the model which results in significant improvement over the best baseline in terms of most metrics (e.g. 14\% and 16\% c-F improvements). This shows that domain ontologies provide additional information that the domain trained embeddings may not contain. While both our methods of incorporating domain ontologies prove to be effective, interpolating domain knowledge directly ($\mathrm{WE}_{Bio}$+dmn) has the edge over retrofitting ($\mathrm{WE}_{Bio}$+rtrft). This is likely due to the direct effect of ontology on the interpolated language model, whereas in retrofitting, the ontology first affects the embeddings and then the context extraction model.

To analyze the performance of our system more closely, we took the context identified by 1 annotator as the candidate and the other 3 as gold standard and evaluated the precision to obtain an estimate of human performance on each citation.
We then divided the citations based on human performance to 4 groups by quartiles. Table \ref{tab:quartile} shows our system's performance on each of these groups. We observe that, when human precision is higher (upper quartiles in the table), our system also performs better and with more confidence (lower std). Therefore, the system errors correlate well with human disagreement on the correct context for the citations.
 Averaged over the 4 annotators for each citation, the mean precision was 56.7\% (note that this translates to our c-P@1 metric). In Table \ref{tab:1}, we observe that our best method (c-P@1 of 56.1\%) is comparable with average human precision score (c-P@1 of 56.7\%) which further demonstrates the effectiveness of our model.

\begin{table}[]
\centering
\renewcommand*{\arraystretch}{0.65}
\small
\caption{Breakdown of our best model's character F-score (c-F) by quartiles of human performance measured by c-P.\vspace{-8pt}}
\label{tab:quartile}
\begin{tabular}{@{}lcccc@{}} \toprule
  Quartiles (c-P)                    & Q1               & Q2  & Q3  & Q4           \\ \midrule
  \begin{tabular}[c]{@{}l@{}} c-F of our model \\(mean $\pm$ stdev.)\end{tabular}     &
  \begin{tabular}[c]{@{}c@{}} 16.14 \\ $\pm$20.20\end{tabular}   &
  \begin{tabular}[c]{@{}c@{}} 25.41 \\ $\pm$7.78\end{tabular}   &
  \begin{tabular}[c]{@{}c@{}} 33.72 \\ $\pm$5.81\end{tabular}   &
  \begin{tabular}[c]{@{}c@{}} 37.50 \\ $\pm$5.93\end{tabular}  \\ \bottomrule
\end{tabular}
\vspace{-6pt}
\end{table}

\vspace{-6pt}
\subsection{External evaluation}
Citation-based summarization can effectively capture various contributions and aspects of the paper by utilizing citation texts \cite{Qazvinian2008scientific}. However; as argued in \autoref{sec:intro}, citation texts do not always accurately reflect the original paper. We show how adding context from the original paper can address this concern, while keeping the benefits of citation-based summarization. Specifically, we compare how using no contextualization, versus various proposed contextualization approaches affect the quality of summarization. We apply the following well-known summarization algorithms on the set of citation texts, and the retrieved citation-contexts: LexRank, LSA-based, SumBasic, and KL-Divergence (For space constraints, we will not explain these approaches here; refer to \cite{nenkova2012survey} for details). We then compare the effect of our proposed contextualization methods using the standard \textsc{Rouge}-N summarization evaluation metrics.

\begin{table}[]
\small
\setlength{\tabcolsep}{2.5pt}
\renewcommand*{\arraystretch}{0.65}
\centering
\caption{Effect of contextualization on summarization. Columns are summarization algorithms and rows show citation contextualization approaches. \textit{No Context} uses only citations without any contextualization. Evaluation metrics are \textsc{Rouge} (\textsc{Rg}) scores. (\textdagger) shows statistically significant improvement over the best baseline performance ($p$<$0.05$).\vspace{-12pt}}
\label{tab:sum-res}
\begin{tabular}{@{}lrrrrrrrr@{}}
\toprule
& \multicolumn{2}{c}{KLSUM}                           & \multicolumn{2}{c}{LexRank}                         & \multicolumn{2}{c}{LSA}                             & \multicolumn{2}{c}{SumBasic}                        \\ \midrule
Method                      & \textsc{Rg}1             & \textsc{Rg}2             & \textsc{Rg}1             & \textsc{Rg}2             & \textsc{Rg}1             & \textsc{Rg}2             & \textsc{Rg}1             & \textsc{Rg}2             \\ \midrule
No Context                  & 36.0                     & 8.3                      & 41.3                     & 10.8                     & 34.7                     & 6.5                      & 38.7                     & 8.7                      \\
VSM \cite{Cohan2015}               & 35.3                     & 7.9                      & 40.0                     & 9.9                      & 33.5                     & 6.2                      & 39.5                     & 9.4                      \\
BM25  & 35.5                     & 8.0                      & 39.8                     & 9.9                      & 33.7                     & 6.0                      & 38.9                     & 8.6                      \\
DESM \cite{mitra2016dual}          & 36.3                     & 8.7                      & 40.2                     & 10.4                     & 32.6                     & 6.5                      & 38.3                     & 7.9                      \\
LMD-LDA \cite{jian2016simple}       & 38.4                     & 9.1                      & 43.1                     & 11.0                     & 37.8                     & 7.6                      & 40.1                     & 8.9                      \\
QR \cite{Cohan2015}                & 39.9                     & 10.2                     & 43.8                     & 11.7                     & 38.9                     & 8.0                      & 40.1                     & 8.6                      \\ \hline
$\mathrm{WE}_\mathrm{Wiki}$                   & 39.7                     & 10.2                     & 42.7                     & 11.8                     & 38.0                     & 8.0                      & 40.2                     & 9.2                      \\
$\mathrm{WE}_\mathrm{Bio}$                    & \I 41.7          & \I 11.7          & \I 45.6          & \I \textbf{13.8} & \I 40.3          & \I 9.1           & \I42.4          & \I \textbf{12.6} \\
$\mathrm{WE}_\mathrm{Bio}$+rtrft           & \I 42.9          & \I 12.2          & \I 46.2          & 11.6                     & \I 40.0          & 8.9                      & \I 41.3          & 9.7                      \\
$\mathrm{WE}_\mathrm{Bio}$+dmn             & \I \textbf{44.0} & \I \textbf{13.4} & \I \textbf{47.3} & \I 13.6          & \I \textbf{42.3} & \I \textbf{10.4} & \I \textbf{44.0} & \I 11.7          \\ \bottomrule
\end{tabular}
\vspace{-6pt}
\end{table}

\textit{Results.} The results of external evaluation are illustrated in Table \ref{tab:sum-res}. The first row (``No context'') shows the performance of each summarization approach solely on the citations without any contextualization. The next 5 rows show the baselines and last 4 rows are our proposed contextualization methods. As shown, effective contextualization positively impacts the generated summaries. For example, our best method is ``$\mathrm{WE}_{Bio}$ + dmn'' which significantly improves the quality of generated summaries in terms of \textsc{Rouge} over the ones without any context.
We observe that two low-performing baseline methods for contextualization according to Table \ref{tab:1} (``VSM'' and ``BM25'') also do not result in any improvements for summarization.
 Therefore, the intrinsic quality of citation contextualization has direct impact on the quality of generated summaries. These results further demonstrate that effective contextualization is helpful for scientific citation-based summarization.

\section{Related work}
Related work has mostly focused on extracting the citation text in the citing article (e.g. \cite{abu2012reference}). In this work, given the citation texts, we focus on extracting its relevant context from the reference paper. Related work have also shown that citation texts can be used in different applications such as summarization \cite{Qazvinian2008scientific,mei2008generating,wan2009whetting,cohan-goharian:2015:EMNLP,jaidka2016overview,cohan2017scientific}. Our proposed model utilizes word embeddings and the domain knowledge. Embeddings have been recently used in general information retrieval models. \citet{vulic2015monolingual} proposed an architecture for learning word embeddings in multilingual settings and used them in document and query representation. \citet{mitra2016dual} proposed dual embedded space model that predicts document aboutness by comparing the centroid of word vectors to query terms. \citet{ganguly2015word} used embeddings to transform term weights in a translation model for retrieval. Their model uses embeddings to expand documents and use co-occurrences for estimation. Unlike these works, we directly use embeddings in estimating the likelihood of query given documents; we furthermore incorporate ways to utilize domain specific knowledge in our model. The most relevant prior work to ours is \cite{Cohan2015} where the authors approached the problem using a vector space model similarity ranking and query reformulations.

\section{Conclusions}
 Citation texts are textual spans in a citing article that explain certain contributions of a reference paper.
 We presented an effective model for contextualizing citation texts (associating them with the appropriate context from the reference paper). We obtained statistically significant improvements in multiple evaluation metrics over several strong baseline, and we matched the human annotators precision. We showed that incorporating embeddings and domain knowledge in the language modeling based retrieval is effective for situations where there are high terminology variations between the source and the target (such as citations and their reference context). Citation contextualization not only can help the readers to better understand the citation texts but also as we demonstrated, they can improve other downstream applications such as scientific document summarization.
  Overall, our results show that citation contextualization enables us to take advantage of the benefits of citation texts, while ensuring accurate dissemination of the claims, ideas and findings of the original referenced paper.


\section*{Acknowledgements}
We thank the three anonymous reviewers for their helpful comments and suggestions. This work was partially supported by National Science Foundation (NSF) through grant CNS-1204347.

\bibliographystyle{ACM-Reference-Format}
\bibliography{sigconf.bib}

\end{document}